\documentclass{article}

\usepackage{arxiv}

\usepackage[utf8]{inputenc} %
\usepackage[T1]{fontenc}    %
\usepackage{hyperref}       %
\usepackage{url}            %
\usepackage{booktabs}       %
\usepackage{amsfonts}       %
\usepackage{amsmath}
\usepackage{nicefrac}       %
\usepackage{microtype}      %
\usepackage{lipsum}		    %
\usepackage{graphicx}
\usepackage{float}
\usepackage{subcaption}
\usepackage{doi}
\usepackage{comment}
\usepackage[square,numbers,sort&compress]{natbib}
\usepackage{wrapfig}

\usepackage{listings}
\usepackage{xcolor} %

\title{CUROCKET: Optimizing ROCKET for GPU }

\author{
    \href{https://orcid.org/0009-0007-0884-0011}{\includegraphics[scale=0.06]{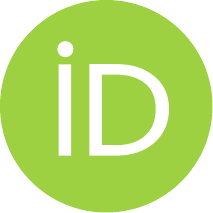}\hspace{1mm}Ole St{\"u}ven}\thanks{All authors are with \textit{Institute of Aircraft Production Technology}, \textit{Hamburg University of Technology}, \textit{Denickestraße 17}, \textit{21073 Hamburg}, \textit{Germany}.}\\
	\texttt{ole.stueven@tuhh.de} \\
    \And
    \href{https://orcid.org/0000-0003-3892-1588}{\includegraphics[scale=0.06]{orcid.pdf}\hspace{1mm}Keno Moenck$^{*}$}\\
	\texttt{keno.moenck@tuhh.de} \\
    \And
    \href{https://orcid.org/0000-0002-9616-3976}{\includegraphics[scale=0.06]{orcid.pdf}\hspace{1mm}Thorsten Sch{\"u}ppstuhl$^{*}$}\\
	\texttt{schueppstuhl@tuhh.de} \\
}

\renewcommand{\shorttitle}{CUROCKET}

\hypersetup{
pdftitle={CUROCKET},
pdfsubject={Optimizing ROCKET for GPU},
pdfauthor={Ole Stueven},
pdfkeywords={pdfkeywords},
}

\begin{document}
\maketitle

\begin{abstract}
ROCKET (RandOm Convolutional KErnel Transform) \cite{dempster_rocket_2020-1} is a feature extraction algorithm created for Time Series Classification (TSC), published in 2019. It applies convolution with randomly generated kernels on a time series, producing features that can be used to train a linear classifier or regressor like Ridge. At the time of publication, ROCKET was on par with the best state-of-the-art algorithms for TSC in terms of accuracy while being significantly less computationally expensive, making ROCKET a compelling algorithm for TSC. This also led to several subsequent versions, further improving accuracy and computational efficiency.
The currently available ROCKET implementations are mostly bound to execution on CPU. However, convolution is a task that can be highly parallelized and is therefore suited to be executed on GPU, which speeds up the computation significantly. A key difficulty arises from the inhomogeneous kernels ROCKET uses, making standard methods for applying convolution on GPU inefficient. In this work, we propose an algorithm that is able to efficiently perform ROCKET on GPU and achieves up to \(11\) times higher computational efficiency per watt than ROCKET on CPU. The code for CUROCKET is available in this repository \url{https://github.com/oleeven/CUROCKET} on github.
\end{abstract}


\setcounter{footnote}{0}%

\section{Introduction}
TSC is a domain within Machine Learning (ML) that focuses on extracting information out of time series. A time series is a list of values of the same variable, measured at different points in time. Despite the name, TSC also includes other applications than classification, e.g., regression or extrinsic regression, which is a form of forecasting. ROCKET itself is independent of the final use, it only extracts features from time series with which a (linear) predictive model can be trained. There are various applications of TSC, e.g. weather, finance, or manufacturing, which is where the motivation for this paper lies, even though the CUROCKET can be applied to other use cases as well. Even though ROCKET offers good scalability compared to most other TSC algorithms, there is room for improvement. The necessity for scalability improvement can be motivated by the discrepancy between the datasets typically used for developing TSC algorithms and data of potential real world use cases. The collection of datasets almost all researchers use for developing TSC algorithms is the UCR archive, which has the advantage of a standardized comparison between algorithms. One disadvantage is that they may not properly reflect real use cases. A major difference is the size of the datasets. When using TSC, e.g., for quality prediction of an automated manufacturing process, the amount of data generated can be enormous compared to the datasets in the UCR archive. A key step to make ROCKET able to better handle these amounts of data, is to implement a more computationally efficient version of it. As the CPU version of ROCKET is already highly optimized, this paper aims to leverage the parallel processing capabilities of GPUs to improve the efficiency of ROCKET.

The rest of this paper is structured as follows: Firstly, the challenge of implementing ROCKET for GPU is pointed out in Section~\ref{sec:rocket_challenges}. Next, the implementation of ROCKET in CUDA is presented in Section~\ref{sec:cuda_rocket}. Afterwards, the performance of CUROCKET is compared to ROCKET in Section~\ref{sec:performance} and its possible variates and limitations are discussed in Section~\ref{sec:variates_and_limitations}. Finally, the results are concluded in Section~\ref{sec:conclusion}.

\section{Challenges of ROCKET for GPU}\label{sec:rocket_challenges}
\subsection{ROCKET} 
\begin{figure}[h]
    \centering
    \includegraphics[width=0.5\textwidth]{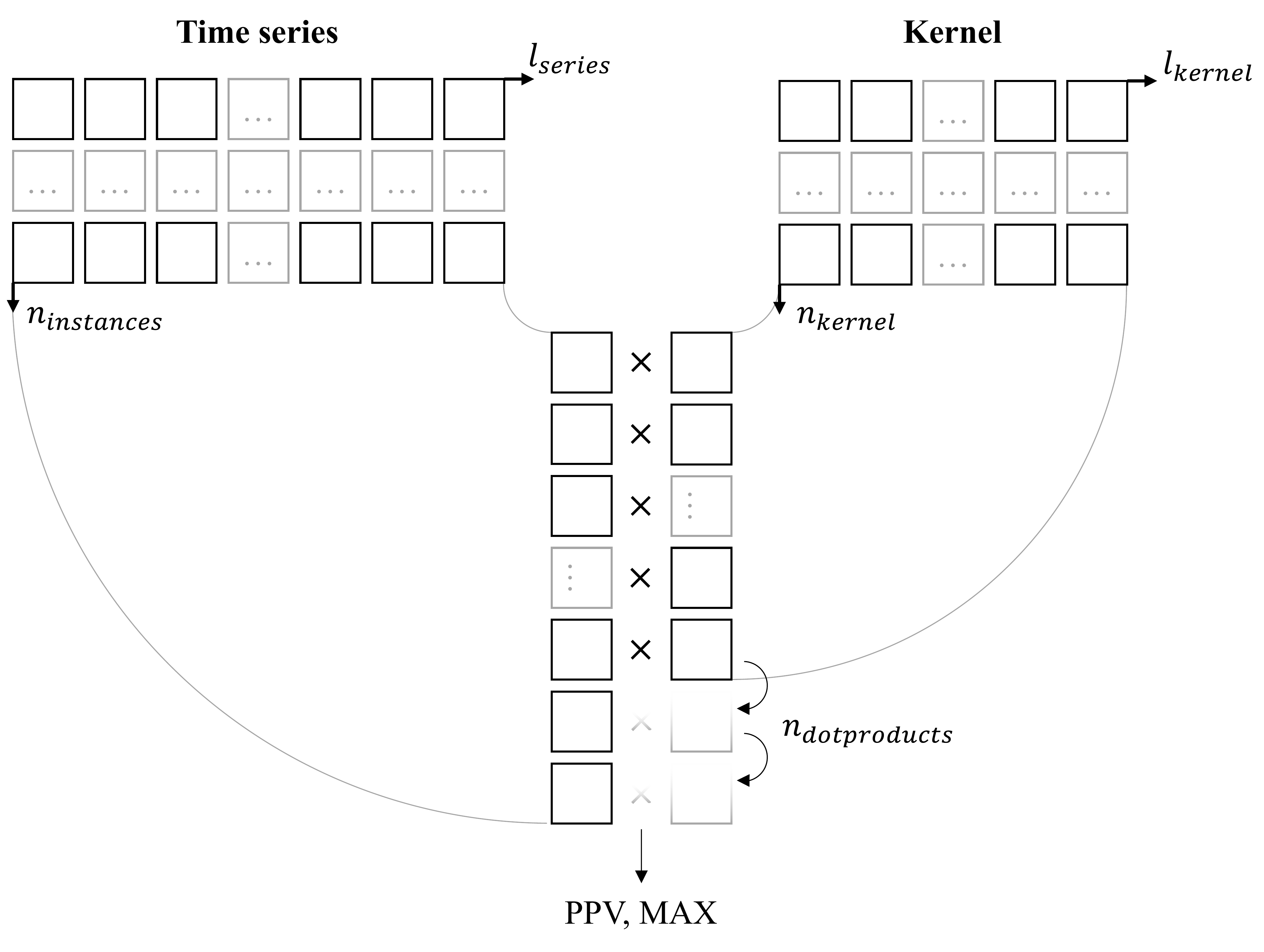}
    \caption{Working principle of the ROCKET algorithm.}
    \label{fig:rocket_principle}
\end{figure}

In the following, we briefly explain ROCKET's principle, which is displayed in Figure~\ref{fig:rocket_principle}.

There are \(n_{kernel}\) kernels randomly generated. For each kernel \textit{length}, \textit{weights}, \textit{bias}, \textit{dilation}, and \textit{padding} are varied. \textit{Length} is selected randomly from \(\{7, 9, 11\}\). \textit{Weights} are sampled from a normal distribution. \textit{Bias} is sampled from a uniform distribution between \(-1\) and \(1\). \textit{Dilation} \(d\) is set to \(2^x\), where \(x\) is drawn from a uniform distribution between \(0\) and \(A\), where
\begin{equation*}
    A=\log_2\frac{l_{series}-1}{l_{kernel}-1},
\end{equation*}
\(l_{series}\) and \(l_{kernel}\) being the length of the time series and kernel respectively.
\textit{Zero padding} is applied with a probability of \(0.5\) so that the kernel is centered on each time point of the series.
Convolution is conducted for each combination of time series and kernel. A single convolution consists of multiple dot products of the kernel with a section of the time series.
Two features are extracted from the convolution result: the maximum value (MAX) and the Proportion of Positive Values (PPV).

\subsection{Insufficiency of standard convolution for ROCKET}
Machine learning libraries for GPUs do not have the ability to handle variable kernel length, padding, and dilation. Kernel length and padding could be circumvented by calling the function three or two times for each. Dilation cannot be handled this way since the set of possible dilations might be huge, depending on the length of the time series. It is inefficient to call a convolution function for each combination of kernel length and dilation. There are two implementation of ROCKET for GPU that both use DL frameworks. The first one is available in tsai \cite{oguiza_tsai_2023} and uses PyTorch. The second one is available in aeon \cite{middlehurst_aeon_2024} and uses TensorFlow. For a random univarate dataset with 1000 instances, a series length of 1000 and 10000 kernels, they are 4.5 and 112.5 times slower than the CPU implementation in sktime. This test was conducted on a machine with a RTX3090 with \(350\,\mathrm{W}\) nominal power as GPU and AMD EPYC 7443P with \(200\,\mathrm{W}\) nominal power as CPU. Corrected for the electrical power, its is 7.8 and 196.7 times slower. Since these existing GPU implementations of ROCKET are slower than those for CPU, a lower-level library must be used to efficiently utilize the GPU for ROCKET.

\section{CUROCKET}\label{sec:cuda_rocket}
\subsection{CUDA programming}
CUDA is an API that enables the use of an NVIDIA GPU for accelerated processing within a computer program.
Analog to the ROCKET section, only the information necessary to understand CUROCKET is given here. For more in-depth information, refer to \cite{noauthor_cuda_nodate}.

\begin{figure}[h]
    \centering
    \includegraphics[width=0.8\textwidth]{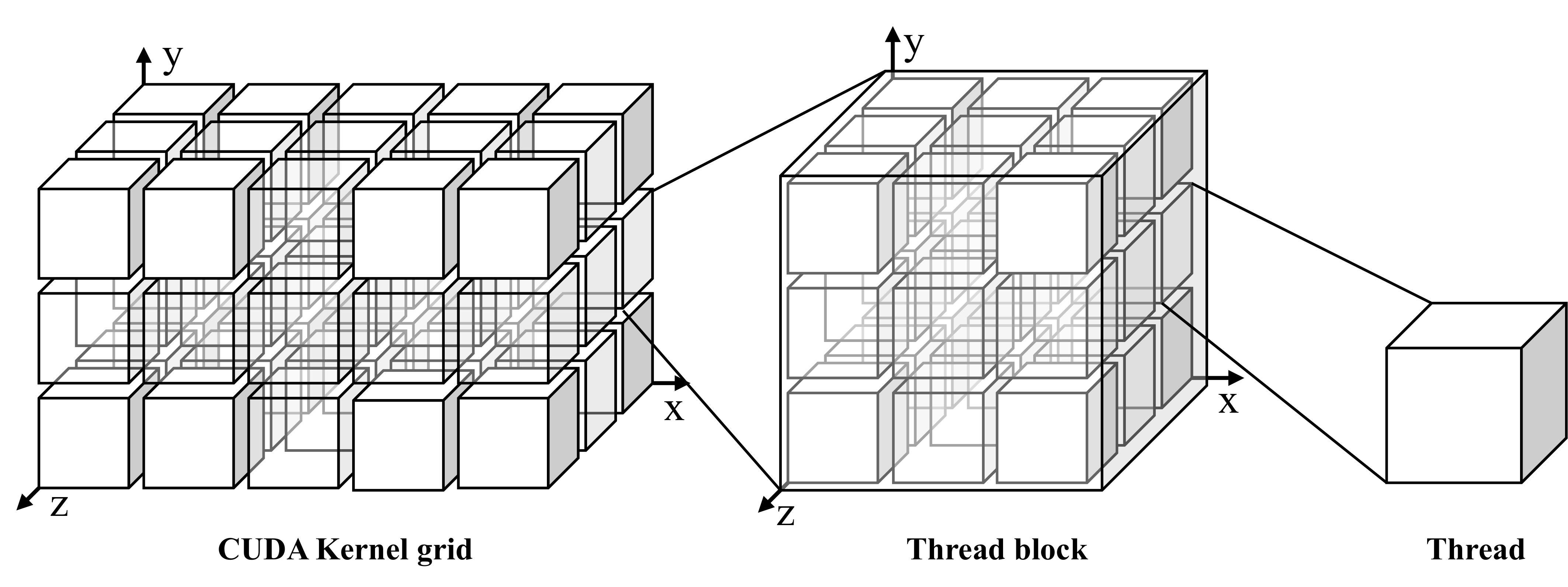}
    \caption{Working principle of CUDA.}
    \label{fig:cuda_principle}
\end{figure}

In Figure~\ref{fig:cuda_principle} the working principle of CUDA is displayed. CUDA takes in a kernel, which defines the instructions for the GPU and the structure of the kernel grid. We call it a CUDA kernel to avoid confusion with a kernel in convolution. The CUDA kernel grid can have up to 3 dimensions, with a maximum shape of \((2^{31}-1, 65535, 65535)\). The grid consists of thread blocks, where each block can contain up to 1024 threads and is executed on a single Streaming Multiprocessor (SM). Each SM has a shared cache for its threads. %

\subsection{Implementing CUROCKET}
\begin{figure}[h]
    \centering
    \includegraphics[width=0.8\textwidth]{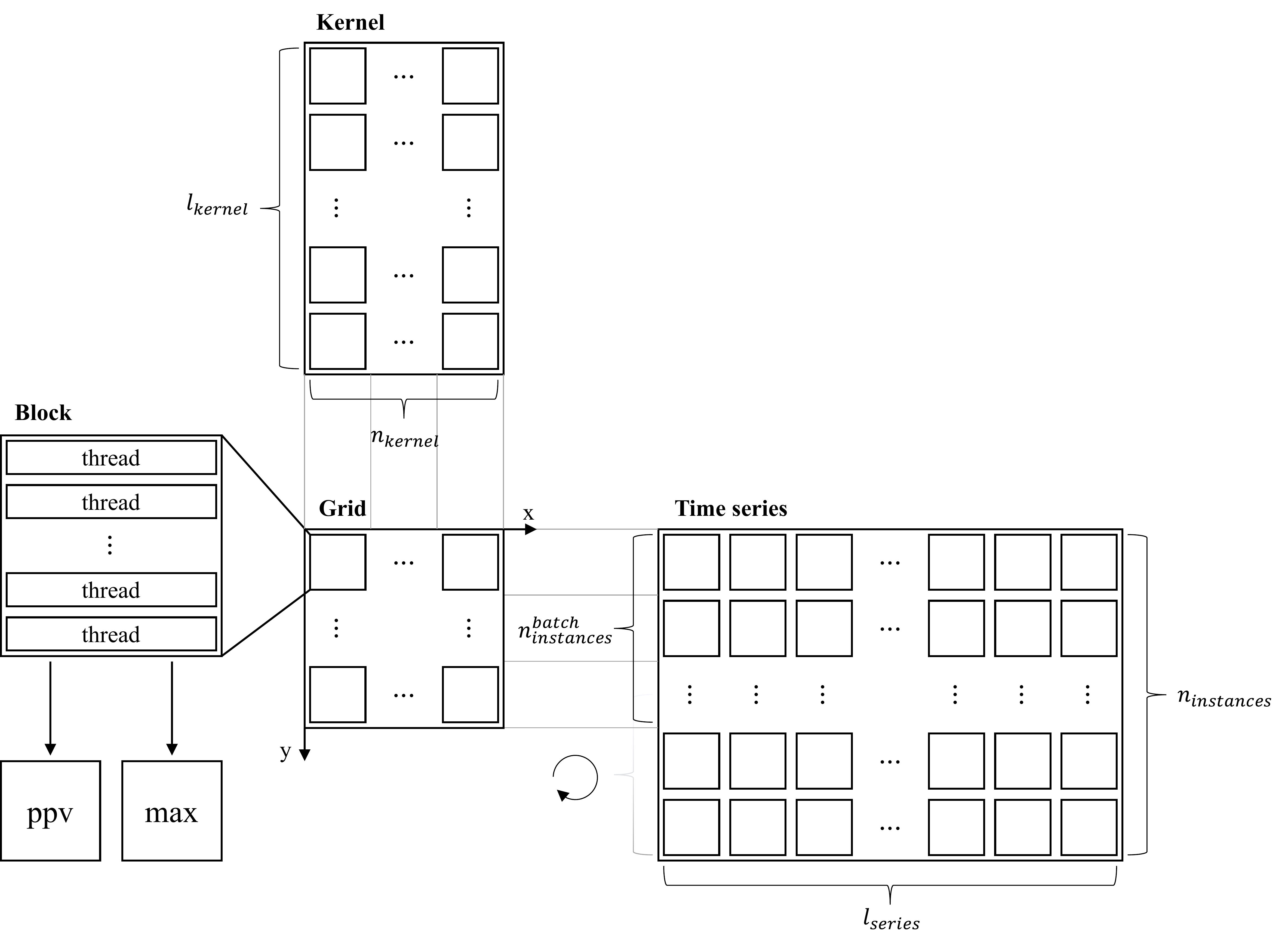} %
    \caption{Working principle of CUROCKET.}
    \label{fig:curocket_principle}
\end{figure}

The main challenge of implementing CUROCKET is to fit the inhomogeneous kernels regarding length, dilation, and padding into the CUDA programming model. There are different approaches to how this can be achieved. The problem can be thought of as aligning the different calculation layers of ROCKET with the layers of CUDA. The working principle of CUROCket is displayed in Figure~\ref{fig:curocket_principle}
Depending on the dataset the number of instances and kernels and the series length can vary immensely. Since the motivation behind CUROCKET is to improve scalability for big datasets, a high number of instances and kernels as well as high series length should be handleable. Each of which might exceed the maximum y dimension of the grid of \(65.535\). Therefore, the number of kernels is mapped to the x dimension of the grid, limiting the maximum number of kernels to \(2^{31}-1\), which is plenty, as in the default setting, ROCKET uses \(10.000\) kernels. The number of instances \(n_{instances}\) is mapped to the y dimension of the grid. To handle more than \(65.535\) instances, the execution of the CUDA kernel is looped in the program.
The reason why kernels and instances are not switched is that the instances will most likely require more memory than the kernels, potentially exceeding the GPU memory.

It is more efficient to always have all kernels on the GPU memory and then cycle through the instances in batches of the minimum of \(65.535\) and the maximum number of instances that can be loaded onto the GPU without exceeding its memory. The next ROCKET layer is the number of dot products, which is mapped to the x dimension of the block. The other two dimensions are not used. Each thread inside the block performs the calculation of one dot product of one kernel and one instance. Since there are only \(1.024\) threads per block, there is a loop within each thread that performs the calculation of another dot product until all dot products for that pair of kernel and instance are done. During the calculation of one dot product, the thread firstly performs the multiplications of each kernel weight with the corresponding timepoint while applying padding and dilation as well as handling edge cases at the start and end of the instance. Bias is added whereupon the PPV counter and current maximum value is updated. Both of which are stored in the block cache. These updates have to be atomic.
After all the threads are finished, the features are written to GPU memory and can be transferred to system memory. The PPV feature has to be divided by the number of calculations per kernel to obtain the real PPV feature.

CUROCKET is implemented in Python.
To access the CUDA API, the Python package CuPy \cite{okuta_cupy_nodate} is used. It allows the use of the GPU with almost identical syntax to NumPy and the execution of custom CUDA kernels written in the CUDA version of C++. Both capabilities are used for CUROCKET. The random generation of the kernels is not implemented on GPU, as it accounts for small a portion of the computing time. To enhance easy of use, it is implemented in an sklearn-compatible format, as is common practice in the TSC research community.

\subsection{Multivariate}
\begin{figure}[h]
    \centering
    \includegraphics[width=0.5\textwidth]{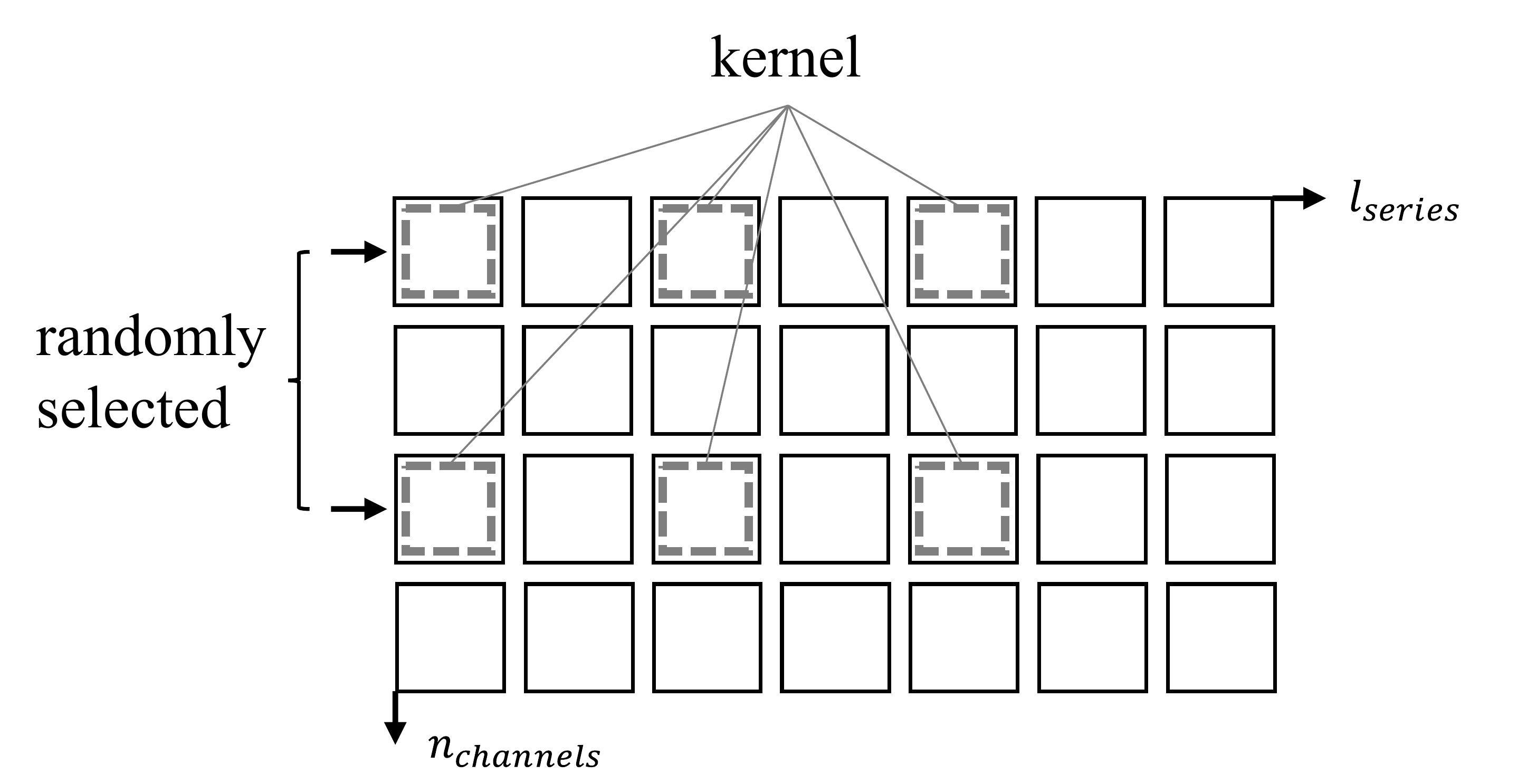} %
    \caption{Principle of multivariate
    ROCKET.}
    \label{fig:rocket_multivar_principle}
\end{figure}

ROCKET was extended to handle multivariate datasets by applying 2d-kernels on the 2d-timeseries. For each kernel the number of channels and which channels it applies to is randomly picked. Figure~\ref{fig:rocket_multivar_principle} shows the principle.

The number of channels and the channel indices are generated by the kernel generation function and are passed to the CUDA kernel via the \lstinline|kernel_data|. The CUDA kernel is modified with a loop that wraps around the dot product loop, calculating the dot product for each selected channel.

\subsection{Multi-GPU}
Generally, the CUROCKET algorithm can easily be split onto multiple GPUs. It can either be split at the instances or the kernels. As with the alignment of ROCKET’s calculation layers to the CUDA layers, it is more efficient to split the instances, as they most likely require more memory than the kernels. Therefore, the instances are split into \(n_{gpu}\) equal parts. For each GPU, a single process is started, that executes CUROCKET. Afterwards, the results from each GPU are merged.

\section{Performance}\label{sec:performance}
\begin{figure}[h]
    \centering
    \includegraphics[width=1.0\textwidth]{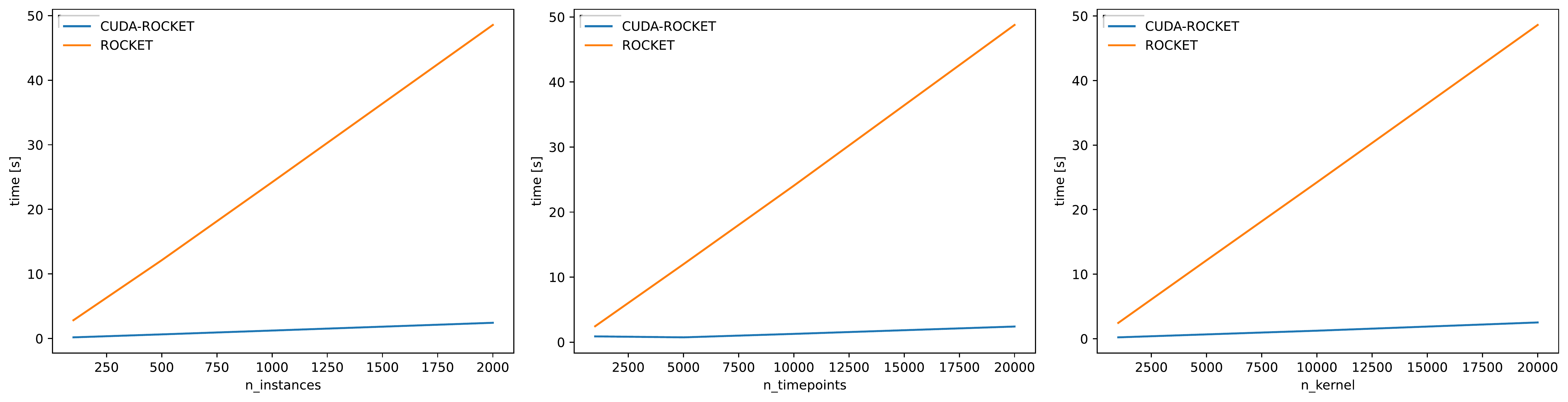} %
    \caption{Comparison of calculating time between CUROCKET and ROCKET, varying the number of instances, series length, and number of kernels from left to right.}
    \label{fig:cuda_rocket_times}
\end{figure}

The performance of CUROCKET was measured in comparison to that of ROCKET, as the existing GPU implementations are slower than ROCKET on CPU. For ROCKET, the implementation in sklearn \cite{pedregosa_scikit-learn_2011} was used. This implementation is highly optimized for CPU, using the numba library \cite{noauthor_numba_nodate} and utilizes all cores. Only the time for feature generation is compared, as the time for kernel generation is the same for both algorithms. Comparing the performance of different algorithms on different devices is not trivial. Therefore, the performance per watt is chosen as a metric. This metric also has disadvantages, as the inherent efficiency of the devices also influences the metric and the exact measurement of power is not possible for devices like CPU and GPU without further modifications. As the devices are under full load during the execution of the algorithms, the nominal power is assumed to be drawn for the full execution of the algorithm.
For the test, an AMD EPYC 7443P with \(200\,\mathrm{W}\) nominal power and a NVIDIA RTX 3090 with \(24\,\mathrm{GB}\) DRAM and \(350\,\mathrm{W}\) nominal power were used.
Test data was randomly generated in the shape \((n_{instances}, l_{series})\). To test if the efficiency changes under different parameters, \(n_{instances}\), \(l_{series}\) and \(n_{kernels}\) was varied. The resulting plots are shown in Figure~\ref{fig:cuda_rocket_times}.

The standard values were set to \(n_{instances} = 1000\), \(l_{series} = 10000\) and \(n_{kernels} = 10000\). \(n_{instances}\) was varied in \(\{100, 500, 1000, 2000\}\), \(l_{series}\) and \(n_{kernels}\) were varied in \(\{1000, 5000, 10000, 20000\}\). The plots show that the calculation time is linear with each parameter. Therefore, the highest pair of values was taken to calculate the speedup of CUROCKET. In the test, the lowest speedup is \(19.3\), which is an efficiency per watt increase of \(11\) of CUROCKET compared to ROCKET.

\section{ROCKET variates and limitations}\label{sec:variates_and_limitations}
The most popular successors of ROCKET are MINIROCKET \cite{dempster_minirocket_2021-1} and MultiRocket \cite{tan_multirocket_2022-1}. The principle of CUROCKET can be applied to MINIROCKET with minor alterations. For MultiRocket, more extensive alterations are necessary because some features of MultiRocket are time-dependent. With CUROCKET, there are only two variables for the features per CUDA kernel that get updated by each thread. For the time-dependent features, the convolution would have to be applied completely before calculating these features. If the convolution result does not fit in the block cache, the current principle does not work. Nevertheless, it is possible to implement MultiRocket for GPU in a similar way, and a speedup compared to CPU is likely. However, one feature of MultiRocket, the Mean of Positive Values (MPV), is not time dependent
and can be added to CUROCKET.

A limitation of CUROCKET is that the PPV feature becomes marginally different from that of ROCKET as the size of the dataset increases, because rounding of float values near zero is different on the GPU than on the CPU. In practice, this doesn’t limit the functionality of CUROCKET, because it just offsets the threshold at which dot products increase the PPV feature, which is equal for every instance.

\section{Conclusion}\label{sec:conclusion}
ROCKET is an established algorithm in TSC, which is already fast compared to most other state-of-the-art TSC algorithms. However, by implementing it for the GPU, further speedup is possible. Due to the inhomogeneity of the kernels, standard convolution functions would be inefficient. Therefore, CUROCKET was implemented with lower-level libraries, achieving a performance per watt increase of \(11\) times compared to CPU. This further increases the scalability of ROCKET and makes it more suitable for real-world use cases.

\section*{Acknowledgements}

\begin{wrapfigure}[3]{r}{0.2\linewidth}
    \raisebox{0pt}[\dimexpr\height-50pt\baselineskip0pt\relax]{
        \includegraphics[width=0.9\linewidth]{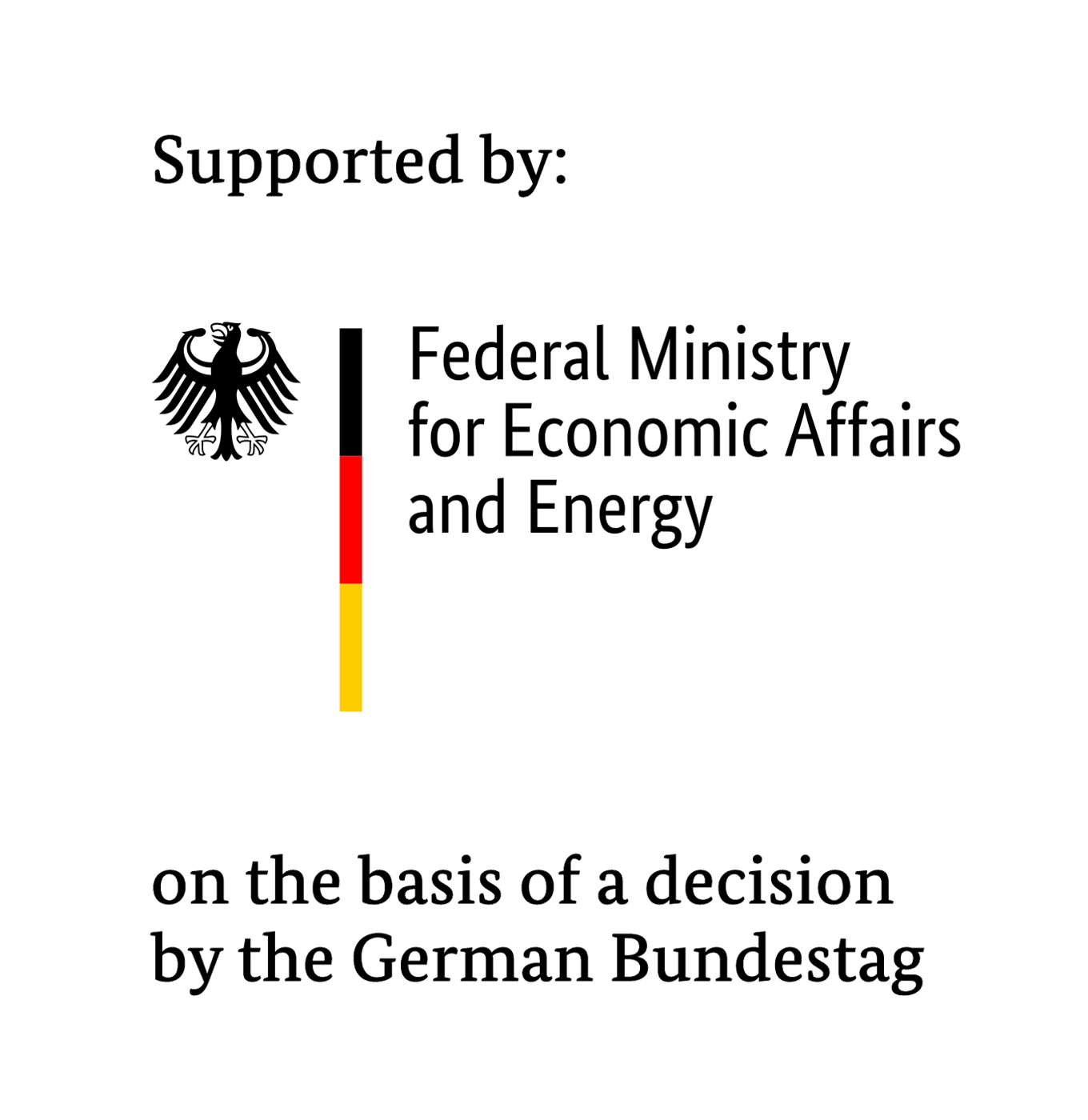}
    }
\end{wrapfigure}

\par

The research leading to these results received funding from the \textit{Federal Ministry for Economic Affairs and Energy (BMWE)} in the project \textit{HoleListic} under the Grant Number 20D2205C as part of the \textit{Federal Aeronautical Research Programme LuFo VI-3}.

\section*{Authors' contributions}
\textbf{Ole Stüven}: Conceptualization, Methodology, Software, Validation, Writing – original draft, Visualization. \textbf{Keno Moenck}: Writing – review and editing. \textbf{Thorsten Schüppstuhl}: Funding acquisition.

\bibliography{references}

\end{document}